\documentclass{article}
\usepackage{ijcai19}

\usepackage{times}
\usepackage{soul}
\usepackage{url}
\usepackage[hidelinks]{hyperref}
\usepackage[utf8]{inputenc}
\usepackage[small]{caption}
\usepackage{graphicx}
\usepackage{amsmath}
\usepackage{booktabs}

\usepackage{leftidx}
\usepackage{color}

\newcommand{\xwc}[1]{\textcolor{red}{\bf [Comments: #1] }}
\usepackage{floatrow}
\usepackage{subcaption}

\title{SPAGAN: Shortest Path Graph Attention Network}

\author{
Yiding Yang$^1$\and
Xinchao Wang$^1$\footnote{Corresponding Author}\and
Mingli Song$^2$\and
Junsong Yuan$^3$\And
Dacheng Tao$^4$\\
\affiliations
$^1$Department of Computer Science, Stevens Institute of Technology\\
$^2$College of Computer Science and Technology, Zhejiang University\\
$^3$Department of Computer Science and Engineering, State University of New York at Buffalo\\
$^4$UBTECH Sydney Artifical Intelligence Centre, University of Sydney\\
\emails
\{yyang99, xwang135\}@stevens.edu,
brooksong@zju.edu.cn,
jsyuan@buffalo.edu,
dacheng.tao@sydney.edu.au.
}

\begin{document}

\maketitle

\begin{abstract}
{Graph} {convolutional} networks~(GCN) have recently demonstrated their potential 
in analyzing non-grid structure data that can be represented as graphs. 
The core idea is to encode the local topology of a graph, 
via convolutions,
into the feature of a center node. 
In this paper, we propose a novel GCN model, which we term as 
Shortest Path Graph Attention Network ({SPAGAN}). 
Unlike conventional GCN models that carry out
{node}-based attentions within each layer,
the proposed SPAGAN conducts \emph{path}-based attention 
that explicitly accounts for the influence of 
a sequence of nodes yielding the minimum cost, or shortest path, 
between the center node and its higher-order neighbors.
SPAGAN therefore allows for a more informative and intact exploration
of the graph structure and further {a} more effective aggregation of 
information from distant neighbors into the center node, as compared to node-based GCN methods.
We test SPAGAN on the downstream classification task
on several standard datasets, 
and achieve performances superior to the state of the art.
Code is publicly available at \url{https://github.com/ihollywhy/SPAGAN}.

\end{abstract}

\section{Introduction}

Convolutional neural networks~(CNNs) have achieved a great success
in many fields especially  computer vision,
where they can automatically learn both low- and high-level 
feature representations of objects within an image, 
thus benefiting the subsequent tasks like classification and detection. 
The input to such CNNs are restricted to grid-like 
structures such as images, in which square-shaped filters can be applied. 

However, there are many other types of data that do not take form of grid structures
and instead represented using graphs. 
For example, in the case of social networks, 
each  person is denoted as a node and each connection as an edge,
and each node may be connected to a different number of edges. 
The features of such  
graphical data 
are therefore encoded in their topological 
structures, which cannot be extracted using 
conventional square or cubic filters tailored for handling 
grid-structured data.

\begin{figure}[t]
\centering
\includegraphics[width=0.60\textwidth]{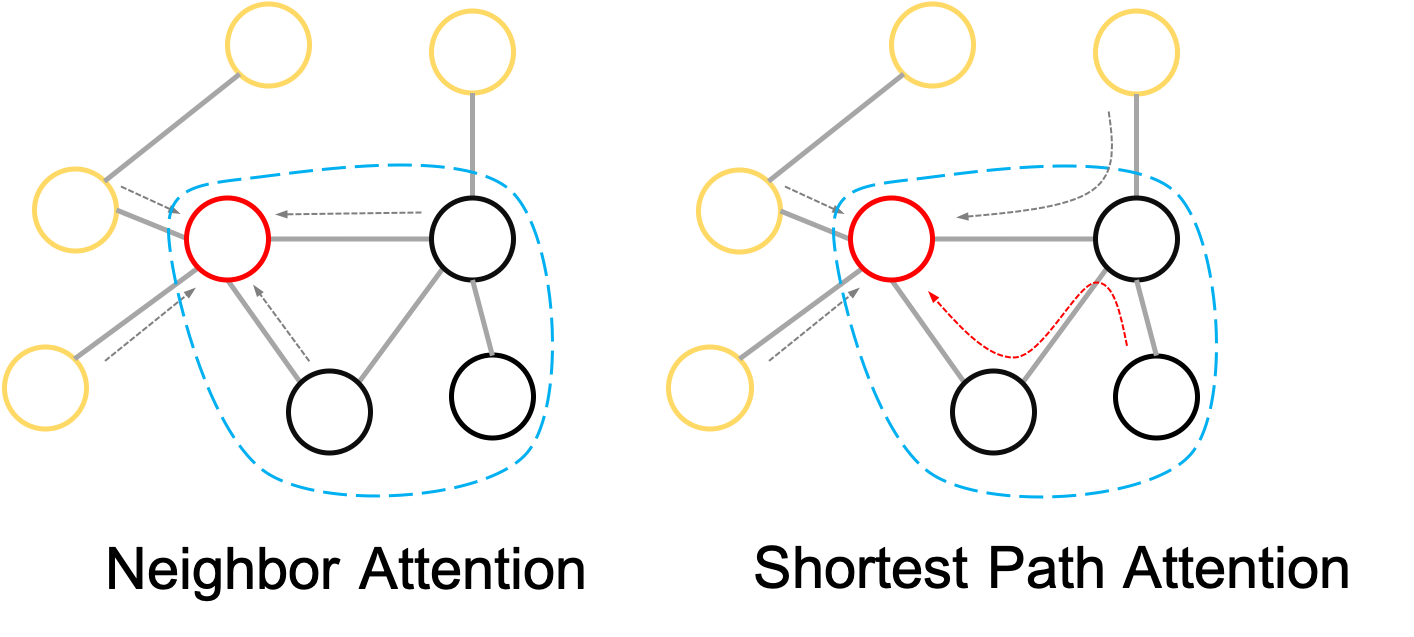}
\caption{
Comparing neighbor attention and shortest path attention. 
Conventional approaches focus on aggregating information from 
first-order neighbors within each layer, shown on the left, 
and often {rely} on stacking multiple layers to reach higher-order neighbors. 
The proposed SPAGAN shown on the right, on the other hand, explicitly conducts shortest-path-based attention that aggregates information from distant neighbors and explores the graph topology, all in a single shot.}
\label{fig:path_aggregation}
\vspace{-0mm}
\end{figure}

To extract features and utilize the contextual information encoded in graphs, 
researchers have explored various possibilities.
\cite{kipf2016semi_GCN} proposes a convolution operator defined in the no-grid domain and build a graph convolutional network (GCN) based on this operator. In GCN, the feature of a center node will be updated by averaging the features of all its immediate neighbors using fixed weights  determined by the Laplacian matrix. 
This feature updating mechanism can be seen as a simplified polynomial filtering of~\cite{ChebNet}, 
which only {considers} first-order neighbors. 
Instead of using all the immediate neighbors, GraphSAGE~\cite{graphsage} suggests using 
only a fraction of them for the sake of computing complexity. 
A uniform sampling strategy is adopted to reconstruct graph and aggregate features. 
Recently, \cite{gat} {introduces} an attention mechanism into the graph convolutional network,
and {proposes} the graph attention network~(GAT) by defining an attention function between each pair of connected nodes.

All the above models, within a single layer, only look at immediate 
or first-order neighboring nodes for aggregating the graph topological information 
and updating features of the center node. 
To account for indirect or higher-order neighbors, one can stack multiple 
{layers} so as to enlarge the {size of} receptive field. However,
results have demonstrated that, by doing so, the performances tend to drop dramatically~\cite{kipf2016semi_GCN}. 
In fact, our experiments even demonstrate that 
{simply}
stacking more layers using of the GAT model will often lead to the failure of convergence.

\begin{figure}[t]
\centering
\includegraphics[width=0.9\textwidth]{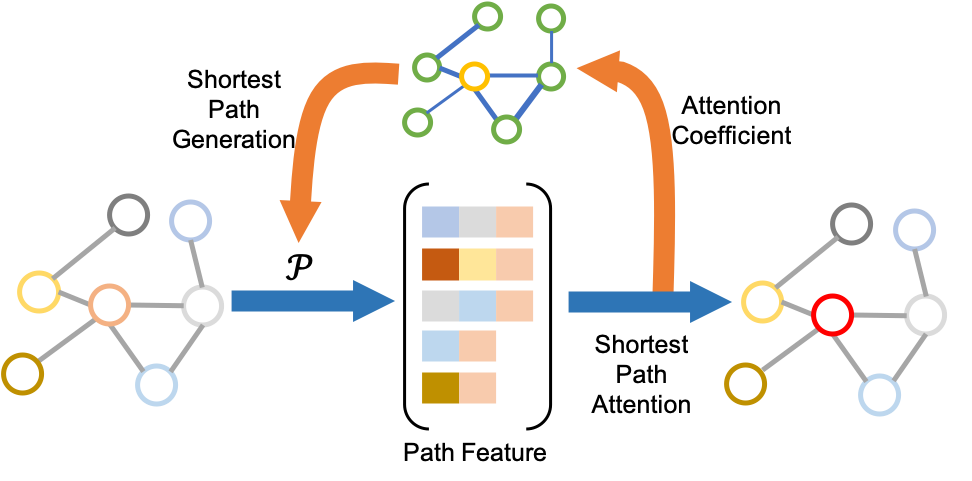}
\vspace{-2mm}
\caption{The overall training workflow of SPAGAN. 
Given a graph, for each center node we compute 
a set of shortest paths of varying lengths to its high-order neighbors
denoted by $\mathcal{P}$, and then extract their features as path features. 
Next, a shortest path attention mechanism is used to compute their attention coefficients with 
{respect}
to the center node. 
The features of each center node  can therefore be updated according to the feature of paths and also their attention coefficients to minimize the loss function. Afterwards, $\mathcal{P}$ will be regenerated according {to} these new attention coefficients and used for the next training phase.
}
\label{fig:framework}
\end{figure}

In this paper, we propose a novel and the first dedicated scheme,
termed Shortest Path Graph Attention Network~(SPAGAN),
that allows us to,
within a single layer, 
utilize path-based high-order attentions to explore the topological information of the graph 
and further  update the features of the center node. 
At the heart of SPAGAN is a mechanism that
finds the shortest paths between a center node and its higher-order neighbors,  then computes a path-to-node attention for updating the node features and coefficients, and iterates the two steps.
In this way, each distant node influences the center node through a path connecting the two with minimum cost,
providing a robust estimation and intact exploration of the graph structure.  

We highlight in  Fig.~\ref{fig:path_aggregation}
the difference between SPAGAN and conventional neighbor-attention approaches. 
Conventional methods look at only immediate neighbors within a single layer. To
propagate information from distant  neighbors, multiple layers would have to be stacked,
often yielding the aforementioned convergence or performance issues.
By contrast,  SPAGAN explicitly explores a sequence of high-order neighbors, 
achieved by shortest paths, to capture the more global graph topology into the path-to-node attetion mechanism.

The training workflow of SPAGAN is depicted in Fig.~\ref{fig:framework}.
For each center node, a set of shortest paths of different lengths
denoted by $\mathcal{P}$, are first computed using the attention coefficients of each pair of nodes that are all initialized to be the same. 
The features of $\mathcal{P}$ will then be generated. 
Afterwards, a path attention mechanism is applied to generate 
the new embedded feature for each node as well as the new attention coefficients. 
The embedded features will be used for computing the loss,
and the attention coefficients, on the other hand, will be used to regenerate 
$\mathcal{P}$ for the next iteration.

Our contribution is therefore a novel high-order graph attention network,
that explicitly conducts path-based attention within each layer, allowing
for an effective and intact encoding of the graphical structure into
the {attention coefficients} and thus  into the features of nodes.
This is achieved by, during training, an iterative scheme that 
computes  shortest paths between a center node and high-order neighbors, 
and estimates the path-based attentions for updating parameters.
We test the proposed SPAGAN on several benchmarks for the downstream node classification task,
where SPAGAN consistently achieves results superior to the state of the art.

\section{Related Work}

In this section, we review related methods on graph convolution and on graph attention mechanism, where the latter one can be considered as a variant of the former. The proposed SPAGAN model falls into the graph attention domain.

\paragraph{Graph Convolutional Networks.} Due to the great success of CNNs, many recent works focus on generating such framework to make it possible to apply on non-grid data structure \cite{gilmer2017neural,MoNet,morris2018weisfeiler,velivckovic2018deep,simonovsky2017dynamic,monti2017geometric,li2018adaptive,zhao2019semantic,magnn}.
\cite{bruna2013spectral} first proposes a convolution operator for graph data by defining a function in spectral domain. It first calculates the eigenvalues and eigenvectors of graph's Laplacian matrix and then applies a function to the eigenvalues. For each layer, the features of node will be a different combination of weighted eigenvectors.
\cite{ChebNet} defines polynomial filters 
on the eigenvalues of the Laplacian matrix to reduce the computational complexity by using the Chebyshev expansion.
\cite{kipf2016semi_GCN} further simplifies the convolution operator by using only first-order neighbors. The adjacent matrix $\mathcal{A}$ is first added self-connections and normalized by the degree matrix $D$. Then, graph convolution operation can be applied by just simple matrix multiplication: $H'=D^{-\frac{1}{2}}\mathcal{A}D^{-\frac{1}{2}}H\Theta$. Under this framework, the convolution operator is defined by parameter $\Theta$ which is not related to the graph structure. Instead of using the full neighbors, GraphSAGE\cite{graphsage} uses a fixed-size set of neighbors for aggregation. This mechanism makes the memory used and running time to be more predictable. \cite{abu2018nrandomwalk} proposes a multi GCN framework by using random walk. They generate multiple adjacent matrices through random walk and feed them to the GCN model.

\paragraph{Graph Attention Networks.} Instead of using fixed aggregation weights, \cite{gat} brings attention mechanism into graph convolutional network and proposes GAT model based on that. 
One of the benefits of attention is the ability to deal with input with variant sizes and make the model focus on parts which are most related to current tasks. In GAT model, the weight for aggregation is no longer equal or fixed. Each edge in the adjacent matrix will be assigned with an attention coefficient that represents how much one node affects the other one. The network is expected to learn to pay more attention to the important neighbors. 
\cite{zhang2018gaan} proposes a gated attention aggregation mechanism which is based on GAT. They argue that not every attention head in GAT model is equally important for the feature embedding process. They propose a model that can learn to assign different weights to different attention head. \cite{geniepath} proposes an adaptive graph convolutional network by combining attention mechanism with LSTM model. They suggest using LSTM model can help to filter the information aggregated from neighbors of variant hops.

\paragraph{Difference to Existing Methods.}
Unlike all the methods described above, the proposed SPAGAN model explicitly conducts
path-based high-order attention that explores the more global graph topology, within in one layer,
to update the network parameters and features of the center nodes. 

\section{Preliminary}
Before introducing our proposed method, we give a brief review of the first-order attention mechanism.

Given a graph convolutional network, let $\mathbf{h}=\{\Vec{h_1},\Vec{h_2},\Vec{h_3},...,\Vec{h_N}\}$ denotes a set of features of $N$ nodes, where $\Vec{h_i}\in \Re^F$ with $F$ being the feature dimension. Also,  let $\mathcal{A}$ denotes the connection relationship, where  $\mathcal{A}_{ij}\in \{0,1\}$ denotes whether there is an edge going from node $i$ to node $j$.
Typically, a linear transformation operation, $W \in \Re^{F' \times F}$, is first applied to the input features individually to transform them into a new space of 
 $F'$ dimension. We write
\begin{equation}
    \leftidx{_{(l)}}{\Vec{h'}}{_j} = \leftidx{_l}{W}{^{(k)}} \leftidx{_{(l-1)}}{\Vec{h}}{_j},
\label{eq:transform}
\end{equation}\noindent
where $\leftidx{_l}{W}{^{(k)}}$ is defined for each layer $l$ and each attention head $k$,
$\leftidx{_{(l-1)}}{\Vec{h}}{_j}$ is the feature of node $j$ in $(l-1)$-th layer, and $\leftidx{_{(l)}}{\Vec{h'}}{_j}$ is the projected feature for node $j$.
Note that, layer zero represents the original feature of input. 

A shared attention function is then applied to those pairs of connected nodes
to obtain the attention coefficients:
\begin{equation}
    \leftidx{_l}{\alpha}{^{(k)}_{ij}}=\frac{exp(\sigma \langle \Vec{a}, \leftidx{_{(l)}}{\Vec{h'}}{_i}\| \leftidx{_{(l)}}{\Vec{h'}}{_j}\rangle)}{\sum\limits_{j\in\mathcal{N}_i}exp(\sigma \langle \Vec{a}, \leftidx{_{(l)}}{\Vec{h'}}{_i}\| \leftidx{_{(l)}}{\Vec{h'}}{_j}\rangle)},
\label{eq:gatatt}
\end{equation}
where $\leftidx{_l}{\alpha}{^{(k)}_{ij}}$ represents the attention coefficient from node~$j$ to node~$i$,  $\langle\cdot, \cdot \rangle$ represents dot product of two vectors, $\mathcal{N}_i$ represents a set of 1-hop neighbors of node $i$, and $\sigma$ can be any non-linear operation. Intuitively, the shared attention function
controls how much the feature of one node affects that of the other node.
By sharing the same vector $\Vec{a}$ for all pairs, 
the number of parameters as well as  the risk of over-fitting can be reduced.

A forward process that updates the node features based on the obtained attentions,
is then carried out:
\begin{equation}
    \leftidx{_{l}}{\Vec{h}}{_i}  = \sigma' \left\{ \displaystyle\mathrel{\mathop{ {\Xi_{k=1}^{K}} }}
    \{ \sum_{j\in \mathcal{N}_i} \leftidx{_l}{\alpha}{^{(k)}_{ij}}  \leftidx{_{(l-1)}}{\Vec{h'}}{_j} \} \right\},
\label{eq:gat}
\end{equation}\noindent
where $K$ is the number of attention head, $\Xi$ denotes the operation that can be
concatenation, max pooling, or mean pooling. For example, \cite{gat} uses concatenation for the middle layers and mean pooling for the last layer.

The above forward process is totally differentiable and can be seen as one graph convolutional layer. Note that the above operation only considers one-hop neighborhoods 
{within one layer}. 
Although stacking multiple such layers can make farther nodes be accessible, it often leads to worse performance and in many cases convergence issues.

\section{Proposed Method}
In this section, we will give more details of the Shortest Path Graph Attention Network (SPAGAN). 
The core idea of SPAGAN is to utilize shortest-path-based attention, which allows us to robustly and effectively explore the more global graph topology, to update the attention functions and the node features within one layer.
The overall structure of SPAGAN is shown in Fig.~\ref{fig:framework} and can be optimized in an iterative manner. The network takes node features $\mathbf{h}$ and the shortest paths
$\mathcal{P}$ to minimize the loss function and thus to update coefficients and features,
based on which the shortest paths are re-computed, and the process iterates.

\subsection{Shortest Path Attention}

\begin{figure}
\centering
\includegraphics[width=0.9\textwidth]{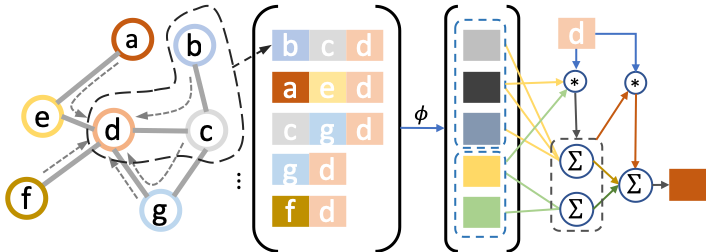}
\caption{Aggregating feature for a center node using shortest path attention. For each center node $d$, shortest paths starting from it can be represented as a set of vectors with different sizes. These vectors will be mapped to the same dimension by the function $\phi$. $*$ denotes the attention operator defined in Eq.~\ref{eq:att}, which will take $d$ as a query to conduct a hierarchical path aggregation.}
\vspace{-2mm}
\label{fig:path_agg}
\end{figure}

We will discuss here how to aggregate the information of shortest paths to the center node, which we call shortest path attention. Different from graph attention network that only focuses on pairwise node attention defined by the adjacent matrix, the proposed SPAGAN approach focuses on 
path-to-node attention.
SPAGAN contains three steps, shortest path generation, path sampling,
and hierarchical path aggregation, for which we give details as follows.

\subsubsection{Shortest Path Generation}

In the initial phase, the network takes the node features $\mathbf{h}$ and the shortest paths $\mathcal{P}$ generated using the same edge weights as input, {to minimize the loss of a specific task, like  cross entropy loss for classification}.
The  attention function, once trained,   generates meaningful edge weights based on  the learned attention coefficients. 
Shortest paths are then computed using  Dijkstra’s algorithm \cite{dijkstra}, where the weights of edges are reversed first and then transformed to positive values using the approach of \cite{suurballe1974disjoint}. Note that the paths here do not include the start node.

To get a robust representation of the weight of an edge, we use the final layer attention coefficients and average those of all attentions:
\begin{equation}
    \mathcal{W}_{ij} = \frac{1}{K}\sum_{k=1}^K \leftidx{_{\Bar{l}} }{\alpha}{^{(k)}_{ij}},
\label{eq:path_weight}
\end{equation}\noindent
where  $\Bar{l}$ denotes the last layer of network, $K$ denotes the number of attention heads for that layer, $\alpha^{(k)}_{ij}$ denotes the attention coefficient from node $j$ to node $i$ in $k$-th head, and $\mathcal{W}_{ij}$ corresponds to the weight of the edge from node $i$ to node $j$.

Let $p_{ij}^c$ denotes the obtained shortest path from node $i$ to node $j$ with length $c$,
and let $\mathcal{P}$ denote the set of all such paths.
Since a path with length $c$ will allow the center node to access nodes up to $c$-hop, 
we can therefore control the {size of} receptive field for a single layer by defining the max value of $c$.  We will give analysis about $c$ in the experimental section. Also, we add the center nodes themselves to $\mathcal{P}$.

\subsubsection{Path Sampling}

For shortest paths of the same length, those with smaller costs are heuristically
more correlated with the center node, and vice versa.
To highlight the contributions of the more correlated paths 
and meanwhile reduce the computational load, for a fixed length $c$,
we sample the top $k$ paths with the lowest costs.
We denote the set of all the sampled paths of length $c$ centerd node $i$ as   $\mathcal{\aleph}_i^c$.
We write

\begin{equation}
    \mathcal{\aleph}_i^c = top_{k}(\mathcal{P}^c),k=\widetilde{deg}_i*r,
    \label{eq:sample}
\end{equation}
where $\mathcal{P}^c$ is the subset of $\mathcal{P}$ that contains all shortest paths of length $c$, and 
$k$ is determined by the degree of the center node, because we want to make the embedded features from paths of different lengths comparable. $\widetilde{deg}_i$ represents the degree of node $i$, and 
$r$ is a hyper-parameter that controls the ratio of the degree of the center node over the number of sampled paths. We will also give analysis about this hyper-parameter in the experimental section.

\subsubsection{Hierarchical Path Aggregation} 

Path aggregation is the core of our model.
We expect the shortest paths of different lengths to encode richer topological information as compared to the first-order neighbors do.
To this end, we propose a hierarchical path aggregation mechanism that 
{focuses on the paths of same length in the first level 
(Eq.~\ref{eq:onepath}) and on different ones in the second level (Eq.~\ref{eq:total}).}

{The hierarchical path aggregation mechanism is shown in Fig.~\ref{fig:path_agg}.
In the first level, given a center node $i$ and its shortest path set $\mathcal{\aleph}_i$, 
we aggregate features respect to each $c$.} 
For the sake of simplicity, we omit $l$ that represents layer in the following discussion. We write
\begin{equation}
    \ell_i^c = \mathop{ {\Xi_{k=1}^{K}} }
    \big \{ \sum_{p^c_{ij}\in \mathcal{\aleph}_i^c} \leftidx{}{\alpha}{^{(k)}_{ij}}  
    \phi( \leftidx{}{p}{^c_{ij}}  ) \big \},
\label{eq:onepath}
\end{equation}
where $\mathcal{\aleph}_i^c$ is a set of shortest paths starting from node $i$ with length of $c$, $\ell_i^c$ is the aggregated feature for node $i$ respect to $\mathcal{\aleph}_i^c$, $K$ is the number of path attention heads that is the same for all $c$, $\Xi$ can be concatenation, max pooling or mean pooling operation. In our implementation, we use $\Xi$ as a concatenation operation for all middle layers and mean pooling operation for the final layer. 
$\phi$ maps paths with variant sizes to fixed ones. 
In our implementation, mean pooling, an order-invariant operator, is used for $\phi$ that computes the average of all nodes' features in a path.
$\leftidx{}{\alpha}{^{(k)}_{ij}}$ is the attention coefficient between node $i$ and path $p^c_{ij}$, and  is taken to be
\begin{equation}
    \leftidx{}{\alpha}{^{(k)}_{ij}}= \mathcal{AT}\Big(\leftidx{_{}}{\Vec{h'}}{_i}, \phi(p^c_{ij})| a_{\alpha}\Big ),
\label{eq:onepathatt}
\end{equation}
where $a_{\alpha}$ is the parameter that defines the attention function $\mathcal{AT}$, and $\Vec{h'}_i$ donates the feature of node $i$ after the linear transformation. Note that when we set $c$ to 2, the generated attention coefficients will be equal to the node attention that can be used to update edges' weights. 

The attention function $\mathcal{AT}$ in Eq.~\ref{eq:onepathatt} is taken to be 
\begin{equation}
    \mathcal{AT}(a, b| {\theta} )
    = \frac{exp(\sigma \langle {\theta}, a\| b\rangle)}{\sum\limits_{b\in\mathcal{\aleph}_a}exp(\sigma \langle {\theta}, a\| b\rangle)},
\label{eq:att}
\end{equation}
where  $\theta$ is its parameter, 
$\mathcal{\aleph}_a$ denotes a set  
defined on $a$, $||$ represents concatenation.
In the case of  Eq.~\ref{eq:onepathatt} or first level aggregation, $\mathcal{\aleph}_a$ denotes $\mathcal{\aleph}_i^c$,
while in the case of Eq.~\ref{eq:second} to be discussed below or second level aggregation, 
$\mathcal{\aleph}_a$ denotes a set of all $\ell_i$.
The output of this function estimates the attention between $a$ and $b$.

{In the second level, we focus on aggregating features of paths with different lengths and applying the attention mechanism to obtain the embedded features for the center node:}
\begin{equation}
    \leftidx{_{}}{\Vec{h}}{_i}  = \sigma \big\{\sum_{c=2}^{C}\leftidx{}{\beta}{_c} \ell_i^c \big\},
\label{eq:total}
\end{equation}
where $\ell_i^c$ is the aggregated feature for node $i$ from paths of length $c$. $C$ is the maximum-allowable path length and $\sigma$ can be any no-linear function. $\leftidx{}{\beta}{_c}$ is the attention coefficient for $\ell_i^c$ and can be obtained from the same attention mechanism with
respect to the center node $i$ using an attention function defined by $a_\beta$:

\begin{equation}
    \leftidx{}{\beta}{_c}= \mathcal{AT}\Big(\leftidx{_{}}{\Vec{h'}}{_i}, \ell_i^c | a_{\beta}\Big ).
\label{eq:second}
\end{equation}

\subsubsection{Iterative Optimization}

The whole network then will be trained in an iterative manner that involves shortest paths $\mathcal{P}$ and all other parameters of the network. In the beginning, the network is trained based on $\mathbf{h}$ and $\mathcal{P}$, where $\mathcal{P}$ is generated using equal edge weights.
When the network converges,
$\mathcal{P}$ will be regenerated according to the attention coefficients in the final layer for the next iteration. In practice, we find two iterations is sufficient to get a good performance, which we will give more discussions in the experimental section.

\subsection{Discussion}
Here, we will discuss some properties of SPAGAN.

\paragraph{Generalized GAT.} SPAGAN can be seen as a generalized GAT, as  it  degenerates to GAT when we set the max value of $c$ to 2 and the sample ratio $r$ to 1.0.

\paragraph{Order Invariance.} Order invariance refers to property that,
the embedded feature for a node is independent of the order of its neighbors. 
It is thus a crucial property since we expect the feature for each node to depend 
only on the structure of the graph but not the order of the nodes. 
All our operations are based on pair-wise attention, which is order invariant.

\paragraph{Node Dependency.} Node dependency refers to that for each node, 
the aggregation process is dependent on its local structure, 
so that every node will have its unique aggregation pattern. 
In SPAGAN, shortest paths starting from center node are expected to explore more global graph structure than 
 first-order mechanisms, 
 which makes the feature aggregation process more node dependent.

\paragraph{Relation to Random Walk.}

The random walk approach~\cite{randomwalk} 
assumes that each node can transmit to its neighbors in equal probability that can be represent as the product of adjacent matrices. 
A $k$ step random walk leads to a matrix of $\mathcal{A}^k$. It suffers from tottering caused by walking through the same edge or  node thus may get stuck at a local structure.
Shortest path, by nature, can alleviate such tottering problem \cite{shortest}.

\section{Experimental Setup}

In this section, we  give the details of the datasets used for validation, the experimental setup, and the compared methods.

\subsection{Experimental Datasets} 

We use three widely used semi-supervised graph datasets,
Cora, Citeseer and Pubmed summarized in Tab.~\ref{tab:datasets}. 
{Following the work of \cite{kipf2016semi_GCN,gat},}
for each dataset, we only use 20 nodes per class for training, 500 nodes for validating and 1000 nodes for testing.
We allow all the algorithms to access the whole graph that includes the node features and edges.
{As done in \cite{kipf2016semi_GCN},}
during training, only part of the nodes are provided with ground truth labels;
at test time, {algorithm will be evaluated by their performance on the test nodes.}

\subsection{Experimental Setup} 

We implement SPAGAN under Pytorch framework \cite{pytorch} and train it with Adam optimizer. We follow a similar learning rate and $L_2$ regularization as GAT~\cite{gat}:
For the Cora dataset, we set the learning rate to 0.005 and the weight of $L_2$ regularization to 0.0005; for the Pubmed dataset, we set the learning rate to 0.01 and the weight of $L_2$ regularization to 0.001. 
For the Citeseer dataset, however, we found that the original setting of parameters under the Pytorch framework yields to  worse performances as compared to those reported in the GAT paper. 
Therefore, we set the learning rate to 0.0085 and the weight of $L_2$ regularization to 0.002 to match the original GAT performance. For all datasets, we set a tolerance window and stop the training process if there is no lower validation loss within it.
We use two graph convolutional layers for all datasets with different attention heads. For the first layer, 8 attention heads for each $c$ is used. 
Each attention head will compute 8 features. Then, an ELU \cite{elu} function is applied. 
In the second layer, we use 8 attention heads for the Pumbed dataset 
and 1 attention head for the other two datasets. The outputs of second layer will be used for classification which have the same dimension as the class number. Dropout is applied to the input of each layer and also to the attention coefficients for each node, with a keep probability of 0.4. 

For all datasets, we set the $r$ to 1.0 which means the number of sampled paths is the same as the degree of each node. 
The max value of $c$ is set to be three for the first layer and two for the last layer for all datasets.
The steps of iteration is set to two. 
For all the datasets, 
we use early stopping based on the cross-entropy loss on validation set. We run the model under the same data split for 10 times and report the mean accuracy as well as standard deviation.
Experiments on the sensitivities of these three hyper-parameters will be provided.

\subsection{Comparison Methods}

We compare our method with several state-of-the-art baselines,
{including MLP, DeepWalk~\cite{perozzi2014deepwalk}, Chebyshev~\cite{ChebNet}, 
GeniePath~\cite{geniepath}, GCN, MoNet~\cite{MoNet} and GAT.}
MLP only uses the node features. 
DeepWalk is a random walk base method. Chebyshev is a spectral method. 
GeniePath uses LSTM model to aggregate features from different layers on a GAT model. 
GCN can be seen as a 
special case of
GAT model where $\alpha_{ij}$ is the same for each pair of nodes. 
MoNet is a first-order spatial method, which uses Gaussian kernel to learn the weight for each neighbors.
GAT is a first-order attention framework which can be seen as a special case of SPAGAN when we set the max length of paths to 2 and the sampling ratio to 1.0. 
We will see the usefulness of path attention when compare to the naive first-order attention model. 
We also compare SPAGAN with GAT model with the same size of receptive field by increasing the number of layers by one. We add skip connections for each layer before ELU activation in three layers GAT model to make it converge.

\begin{table}
\centering
\begin{tabular}{lccc}  
\toprule
  &  Cora & Citeseer & Pubmed \\
\midrule
\#Nodes       & 2,708  & 3,327 & 19,717 \\
\#Edges       & 5,429  & 4,732 & 44,338  \\
\#Features    & 1,433  & 3,703 & 500 \\
\#Classes     & 7  & 6  & 3 \\     
\bottomrule
\end{tabular}
\vspace{-2mm}
\caption{Summary of the datasets used in our experiments. For each dataset,
we use only 20 nodes per class  for training, 500 nodes for validation, and 1000 nodes for testing.}
\label{tab:datasets}
\end{table}

\begin{table}
\centering
\begin{tabular}{lccc}  
\toprule
 Method &  Cora & Citeseer & Pubmed\\
\midrule
MLP & 55.1\% & 46.5\% & 71.4\% \\
DeepWalk & 67.2\% & 43.2\% & 65.3\%\\
Chebyshev & 81.2\% &  69.8\% & 74.4\% \\
GeniePath & - & - & 78.5\% \\
GCN & 81.5\% & 70.3\% & 79.0\%\\
MoNet & 81.7$\pm$0.5\% & - & 78.8$\pm$0.3\%\\
GAT & 83.0$\pm$0.7\% & 72.5$\pm$0.7\% & 79.0$\pm$0.3\%\\
GAT$^3$ &  81.2$\pm$0.6\%    &  68.8$\pm$0.7\%    & 78.5$\pm$0.4\%\\
\midrule
SPAGAN & \textbf{83.6$\pm$0.5\%}
& \textbf{73.0$\pm$0.4\%} & \textbf{79.6$\pm$0.4\%}\\
\bottomrule
\vspace{-2mm}
\end{tabular}
\caption{Summary of results in terms of classification accuracy. GAT$^3$ denotes
the three-layer GAT model with skip connections (GAT model fails to converge when we simply stack three layers without skip connections).} 
\vspace{-4mm}
\label{tab:results}
\end{table}

\section{Experimental Result}
As we focus on node classification, 
we use classification accuracy to evaluate all methods. 
The results of our comparative experiments are summarized in 
Tab.~\ref{tab:results},
where we observe SPAGAN consistently achieves the best performance on all three datasets. 
This implies that SPAGAN, by exploring high-order neighbors and graph structure
within one layer, indeed leads to more discriminant node features.

\begin{figure}[t]
\centering
\vspace{-2mm}
\includegraphics[width=0.80\textwidth]{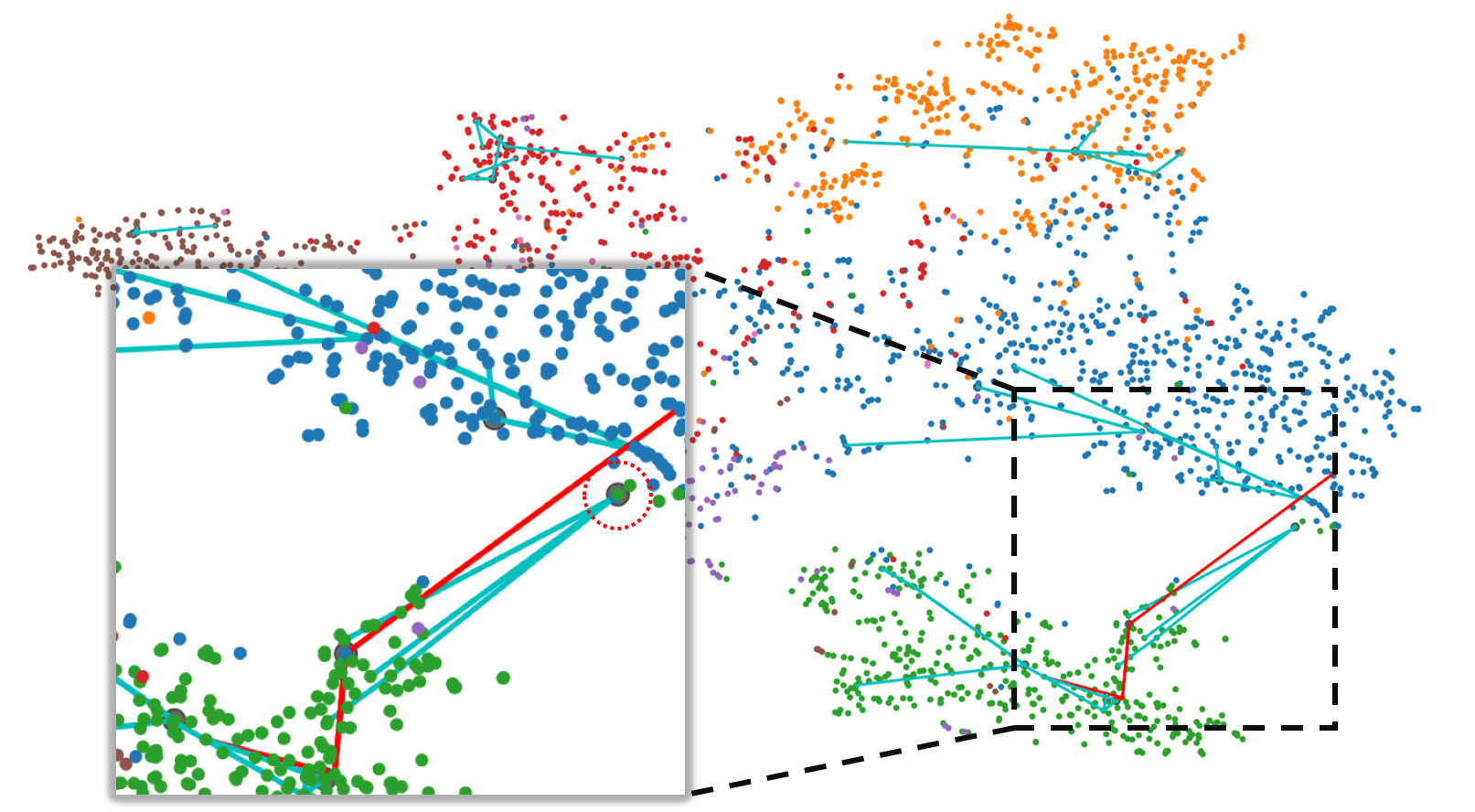}
\caption{Visualization of features obtained from hidden layer of a trained SPAGAN model on the Cora dataset and paths starting from 10 random nodes (with black background). Color of nodes represents their labels. Paths starting from correctly classified nodes are colored with cyan, otherwise with red.}
\vspace{-2mm}
\label{fig:vis}
\end{figure}

In addition,  we provide a visualization on the learned embedded features as well as the shortest paths. The features obtained from the hidden layers of a trained model on the Cora dataset are projected using t-SNE algorithm \cite{t-sne}. 
As shown in Fig.~\ref{fig:vis}, we randomly sample 10 nodes 
and visualize four paths per node.  Nodes with black surroundings 
indicate the starts of paths. 
For the 
correctly classified nodes, the paths are colored with cyan,
otherwise the paths are colored with red.
As can be observed, 
the shortest paths of most correctly classified nodes tend to 
reside within the same cluster. 
Another interesting observation is that, 
for nodes that lie in the wrong cluster, 
shortest paths can still lead them back to the right cluster
and further produce the correct classification results.
For example, in the zoomed region, 
the node surrounded by the red dotted circle lies in blue cluster.
However, since it is connected to shortest paths leading back to the green cluster,
our network eventually predicts the correct label, green, to this node.

\subsection{Sensitivity Analysis}
There are three hyper-parameters in our model: the sampling rate of path $r$, the maximum depth of path $C$, and the number of iteration $Iter$. We change each of them to analyze the sensitive of our model respect to them. 
The results are shown in Tab.~\ref{tab:sensitive}. 
All these results are obtained from the Cora dataset. For the sampling ratio, we fix $Iter$ to be 2 and $C$ to 3 (for sampling ratio 0.5, we adopt a max operator over the number of sampled paths and one, to ensure that at least one path is selected); the best performance is obtained in 1.0, meaning that we use the same number of paths as the degree of each node. Then we fix the sampling ratio to 1.0 and change the number of iterations. Results show that two iterations is sufficient for a good performance. We also change the max depth of paths from 2 to 5. The best performance is obtained at 3.

\begin{table}

\begin{subtable}{1.0\textwidth}
\caption{Results with different sampling ratios.}
\scalebox{0.92}{
\begin{tabular}{ccccc} 
\toprule[1pt]
r &  0.5 & 1.0 & 1.5 & 2.0\\
\midrule
Acc &  83.4$\pm$0.3\% & 83.6$\pm$0.5\% & 83.4$\pm$0.6\% & 83.3$\pm$0.5\%\\
\bottomrule[1pt]
\end{tabular}}
\label{tab:sensitive_a}
\end{subtable}

\begin{subtable}{1.0\textwidth}
\caption{Results with different steps of iterations.}
\scalebox{0.75}{
\begin{tabular}{cccccc}  
\toprule[1pt]
$Iter$ &  1 & 2 & 3 & 4 & 5\\
\midrule
Acc & 83.0$\pm$0.7\% &  83.6$\pm$0.5\% & 83.6$\pm$0.6\% & 83.6$\pm$0.6\% & 83.5$\pm$0.6\% \\
\bottomrule[1pt]
\end{tabular}}
\label{tab:sensitive_b}
\end{subtable}

\begin{subtable}{1.0\textwidth}
\caption{Results with different lengths of paths.}
\scalebox{0.92}{
\begin{tabular}{ccccc}  
\toprule[1pt]
C &  2 & 3 & 4 & 5\\
\midrule
Acc &  83.0$\pm$0.7\% & 83.6$\pm$0.5\% & 83.5$\pm$0.7\% & 83.1$\pm$0.4\%\\
\bottomrule[1pt]
\end{tabular}}
\label{tab:sensitive_c}
\end{subtable}
\vspace{-2mm}
\caption{Sensitivity analysis of hyper-parameters. 
All these results are obtained on the Cora dataset.
}
\label{tab:sensitive}
\end{table}

\begin{table}
\centering
\begin{tabular}{lccc}  
\toprule
Method &  Cora & Citeseer & Pubmed \\
\midrule
GAT & 83.0$\pm$0.7\% & 72.5$\pm$0.7\% & 79.0$\pm$0.3\% \\
\midrule
SPAGAN\_Fix  & 83.2$\pm$0.4\% & 72.4$\pm$0.4\% & 79.2$\pm$0.4\% \\
\midrule
SPAGAN   & 83.6$\pm$0.5\%    & 73.0$\pm$0.4\% & 79.6$\pm$0.4\% \\
\bottomrule
\end{tabular}
\vspace{-2mm}
\caption{Classification accuracy of SPAGAN with and without the hierarchical path aggregation. SPAGAN\_Fix refers to the model where 
we average the features coming from paths with different lengths
(i.e., removing $\beta$ from Eq.~\ref{eq:total}).}
\label{tab:ablation}
\end{table}

\subsection{Ablation Study}
We conduct ablation study to validate the proposed hierarchical path attention mechanism described by Eq.~\ref{eq:second}. We compare the results between full SPAGAN model that has  learnable attention coefficients for paths of different lengths, and SPAGAN model without it, which we call SPAGAN\_Fix. 
The results are shown in Tab.~\ref{tab:ablation}, 
where the full model of SPAGAN improvement over. 
We can thus conclude that hierarchical attention mechanism indeed plays
an important role in the feature aggregation process.

\subsection{Computing Complexity}

Many real-world graphs are sparse.
For example, the average degree of the Cora dataset is only 2.
A sparse graph can be represented by values and indexes in $O(E)$ space complexity where $E$ is the number of edges in the graph. The computed path matrix $\mathcal{P}$ is cutoff by $C$ and 
further reduced by sampling. In our experiments, the space complexity of path attention layer is one or two times larger than that of the first-order one, which is very friendly for GPU memory. Take Cora dataset as an example, GAT model takes about 600M GPU memory while our SPAGAN model only increases it by about 200M.

As for the running time, the shortest path algorithm we adopted, Dijkstra,
runs at a time complexity of $O(VElogV)$ where $V$ is the number of nodes in a graph.
The path attention mechanism is implemented using sparse operator~\cite{scatter} according to 
indices and values of $\mathcal{P}$, which allows us to make full use of the GPU computing ability. 
The running time of one epoch with path attention on Pubmed dataset is 0.1s on a Nvidia 1080Ti GPU.

\section{Conclusion}
We propose in this paper an effective and the first dedicated graph attention network, termed Shortest Path Graph Attention Network (SPAGAN), that allows us to explore high-order path-based attentions within one layer.  This is achieved by computing the shortest paths of different lengths between a center node and its distant neighbors, then carrying path-to-node attention for updating the node features and attention coefficients, and iterates. 
We test SPAGAN on several benchmarks, and demonstrate that it yields results superior to the state of the art on the downstream node-classification task.

\newpage

\bibliographystyle{named}
\bibliography{ijcai19}

\begin{thebibliography}{}

\bibitem[\protect\citeauthoryear{Abu-El-Haija \bgroup \em et al.\egroup
  }{2018}]{abu2018nrandomwalk}
Sami Abu-El-Haija, Amol Kapoor, Bryan Perozzi, and Joonseok Lee.
\newblock N-gcn: Multi-scale graph convolution for semi-supervised node
  classification.
\newblock {\em arXiv preprint arXiv:1802.08888}, 2018.

\bibitem[\protect\citeauthoryear{Borgwardt and Kriegel}{2005}]{shortest}
Karsten~M Borgwardt and Hans-Peter Kriegel.
\newblock Shortest-path kernels on graphs.
\newblock In {\em IEEE international conference on data mining}, 2005.

\bibitem[\protect\citeauthoryear{Bruna \bgroup \em et al.\egroup
  }{2014}]{bruna2013spectral}
Joan Bruna, Wojciech Zaremba, Arthur Szlam, and Yann Lecun.
\newblock Spectral networks and locally connected networks on graphs.
\newblock In {\em International Conference on Learning Representations}, 2014.

\bibitem[\protect\citeauthoryear{Clevert \bgroup \em et al.\egroup
  }{2015}]{elu}
Djork-Arn{\'e} Clevert, Thomas Unterthiner, and Sepp Hochreiter.
\newblock Fast and accurate deep network learning by exponential linear units
  (elus).
\newblock {\em arXiv preprint arXiv:1511.07289}, 2015.

\bibitem[\protect\citeauthoryear{Defferrard \bgroup \em et al.\egroup
  }{2016}]{ChebNet}
Micha{\"e}l Defferrard, Xavier Bresson, and Pierre Vandergheynst.
\newblock Convolutional neural networks on graphs with fast localized spectral
  filtering.
\newblock In {\em Advances in Neural Information Processing Systems}, pages
  3844--3852, 2016.

\bibitem[\protect\citeauthoryear{Dijkstra}{1959}]{dijkstra}
Edsger~W Dijkstra.
\newblock A note on two problems in connexion with graphs.
\newblock {\em Numerische mathematik}, 1(1):269--271, 1959.

\bibitem[\protect\citeauthoryear{Fey \bgroup \em et al.\egroup
  }{2018}]{scatter}
Matthias Fey, Jan~Eric Lenssen, Frank Weichert, and Heinrich M{\"u}ller.
\newblock {SplineCNN}: Fast geometric deep learning with continuous {B}-spline
  kernels.
\newblock In {\em IEEE Conference on Computer Vision and Pattern Recognition},
  2018.

\bibitem[\protect\citeauthoryear{Gilmer \bgroup \em et al.\egroup
  }{2017}]{gilmer2017neural}
Justin Gilmer, Samuel~S Schoenholz, Patrick~F Riley, Oriol Vinyals, and
  George~E Dahl.
\newblock Neural message passing for quantum chemistry.
\newblock In {\em International Conference on Machine Learning}, pages
  1263--1272. JMLR. org, 2017.

\bibitem[\protect\citeauthoryear{Hamilton \bgroup \em et al.\egroup
  }{2017}]{graphsage}
Will Hamilton, Zhitao Ying, and Jure Leskovec.
\newblock Inductive representation learning on large graphs.
\newblock In {\em Advances in Neural Information Processing Systems}, pages
  1024--1034, 2017.

\bibitem[\protect\citeauthoryear{Kashima \bgroup \em et al.\egroup
  }{2003}]{randomwalk}
Hisashi Kashima, Koji Tsuda, and Akihiro Inokuchi.
\newblock Marginalized kernels between labeled graphs.
\newblock In {\em International conference on machine learning}, pages
  321--328, 2003.

\bibitem[\protect\citeauthoryear{Kipf and Welling}{2016}]{kipf2016semi_GCN}
Thomas~N Kipf and Max Welling.
\newblock Semi-supervised classification with graph convolutional networks.
\newblock {\em arXiv preprint arXiv:1609.02907}, 2016.

\bibitem[\protect\citeauthoryear{Li \bgroup \em et al.\egroup
  }{2018}]{li2018adaptive}
Ruoyu Li, Sheng Wang, Feiyun Zhu, and Junzhou Huang.
\newblock Adaptive graph convolutional neural networks.
\newblock In {\em AAAI Conference on Artificial Intelligence}, 2018.

\bibitem[\protect\citeauthoryear{Liu \bgroup \em et al.\egroup
  }{2018}]{geniepath}
Ziqi Liu, Chaochao Chen, Longfei Li, Jun Zhou, Xiaolong Li, Le~Song, and Yuan
  Qi.
\newblock Geniepath: Graph neural networks with adaptive receptive paths.
\newblock {\em arXiv preprint arXiv:1802.00910}, 2018.

\bibitem[\protect\citeauthoryear{Maaten and Hinton}{2008}]{t-sne}
Laurens van~der Maaten and Geoffrey Hinton.
\newblock Visualizing data using t-sne.
\newblock {\em Journal of machine learning research}, 9(Nov):2579--2605, 2008.

\bibitem[\protect\citeauthoryear{Monti \bgroup \em et al.\egroup
  }{2017a}]{MoNet}
Federico Monti, Davide Boscaini, Jonathan Masci, Emanuele Rodola, Jan Svoboda,
  and Michael~M Bronstein.
\newblock Geometric deep learning on graphs and manifolds using mixture model
  cnns.
\newblock In {\em IEEE Conference on Computer Vision and Pattern Recognition},
  volume~1, page~3, 2017.

\bibitem[\protect\citeauthoryear{Monti \bgroup \em et al.\egroup
  }{2017b}]{monti2017geometric}
Federico Monti, Davide Boscaini, Jonathan Masci, Emanuele Rodola, Jan Svoboda,
  and Michael~M Bronstein.
\newblock Geometric deep learning on graphs and manifolds using mixture model
  cnns.
\newblock In {\em IEEE Conference on Computer Vision and Pattern Recognition},
  pages 5115--5124, 2017.

\bibitem[\protect\citeauthoryear{Morris \bgroup \em et al.\egroup
  }{2018}]{morris2018weisfeiler}
Christopher Morris, Martin Ritzert, Matthias Fey, William~L Hamilton, Jan~Eric
  Lenssen, Gaurav Rattan, and Martin Grohe.
\newblock Weisfeiler and leman go neural: Higher-order graph neural networks.
\newblock {\em arXiv preprint arXiv:1810.02244}, 2018.

\bibitem[\protect\citeauthoryear{Paszke \bgroup \em et al.\egroup
  }{2017}]{pytorch}
Adam Paszke, Sam Gross, Soumith Chintala, Gregory Chanan, Edward Yang, Zachary
  DeVito, Zeming Lin, Alban Desmaison, Luca Antiga, and Adam Lerer.
\newblock Automatic differentiation in pytorch.
\newblock 2017.

\bibitem[\protect\citeauthoryear{Peng \bgroup \em et al.\egroup }{2018}]{magnn}
Hao Peng, Jianxin Li, Qiran Gong, Yuanxing Ning, and Lihong Wang.
\newblock Graph convolutional neural networks via motif-based attention.
\newblock {\em arXiv preprint arXiv:1811.08270}, 2018.

\bibitem[\protect\citeauthoryear{Perozzi \bgroup \em et al.\egroup
  }{2014}]{perozzi2014deepwalk}
Bryan Perozzi, Rami Al-Rfou, and Steven Skiena.
\newblock Deepwalk: Online learning of social representations.
\newblock In {\em ACM SIGKDD international conference on Knowledge discovery
  and data mining}, 2014.

\bibitem[\protect\citeauthoryear{Simonovsky and
  Komodakis}{2017}]{simonovsky2017dynamic}
Martin Simonovsky and Nikos Komodakis.
\newblock Dynamic edge conditioned filters in convolutional neural networks on
  graphs.
\newblock In {\em IEEE Conference on Computer Vision and Pattern Recognition},
  2017.

\bibitem[\protect\citeauthoryear{Suurballe}{1974}]{suurballe1974disjoint}
JW~Suurballe.
\newblock Disjoint paths in a network.
\newblock {\em Networks}, 4(2):125--145, 1974.

\bibitem[\protect\citeauthoryear{Veli{\v{c}}kovi{\'c} \bgroup \em et al.\egroup
  }{2018}]{velivckovic2018deep}
Petar Veli{\v{c}}kovi{\'c}, William Fedus, William~L Hamilton, Pietro Li{\`o},
  Yoshua Bengio, and R~Devon Hjelm.
\newblock Deep graph infomax.
\newblock {\em arXiv preprint arXiv:1809.10341}, 2018.

\bibitem[\protect\citeauthoryear{Veličković \bgroup \em et al.\egroup
  }{2018}]{gat}
Petar Veličković, Guillem Cucurull, Arantxa Casanova, Adriana Romero, Pietro
  Liò, and Yoshua Bengio.
\newblock Graph attention networks.
\newblock In {\em International Conference on Learning Representations}, 2018.

\bibitem[\protect\citeauthoryear{Zhang \bgroup \em et al.\egroup
  }{2018}]{zhang2018gaan}
Jiani Zhang, Xingjian Shi, Junyuan Xie, Hao Ma, Irwin King, and Dit-Yan Yeung.
\newblock Gaan: Gated attention networks for learning on large and
  spatiotemporal graphs.
\newblock {\em arXiv preprint arXiv:1803.07294}, 2018.

\bibitem[\protect\citeauthoryear{Zhao \bgroup \em et al.\egroup
  }{2019}]{zhao2019semantic}
Long Zhao, Xi~Peng, Yu~Tian, Mubbasir Kapadia, and Dimitris~N Metaxas.
\newblock Semantic graph convolutional networks for 3d human pose regression.
\newblock {\em arXiv preprint arXiv:1904.03345}, 2019.

\end{thebibliography}

\end{document}